\documentclass[10pt,twocolumn,letterpaper]{article}
\usepackage{cvpr}
\usepackage{times}
\usepackage{epsfig}
\usepackage{graphicx}
\usepackage{amsmath}
\usepackage{amssymb}
\usepackage{comment}
\usepackage{lipsum}

\usepackage{siunitx}        
\usepackage{xcolor}
\usepackage{amsthm}

\usepackage{amsmath,amsfonts,bm}

\def\eqref#1{equation~\ref{#1}}

\def\1{\bm{1}}

\def\vx{{\bm{x}}}
\def\vy{{\bm{y}}}

\def\evy{{y}}

\DeclareMathAlphabet{\mathsfit}{\encodingdefault}{\sfdefault}{m}{sl}
\SetMathAlphabet{\mathsfit}{bold}{\encodingdefault}{\sfdefault}{bx}{n}

\usepackage[pagebackref=true,breaklinks=true,letterpaper=true,colorlinks,bookmarks=false]{hyperref}

\cvprfinalcopy 

\def\cvprPaperID{6346} 

\ifcvprfinal\fi
\begin{document}

\title{Differentiable Adaptive Computation Time for Visual Reasoning}

\author{Cristobal Eyzaguirre and Alvaro Soto\\
Pontificia Universidad Catolica de Chile\\
{\tt\small ceyzaguirre4@uc.cl,     \tt\small asoto@ing.puc.cl}
}

\maketitle

\begin{abstract}
  This paper presents a novel attention-based algorithm for achieving adaptive computation called DACT, which, unlike existing ones, is end-to-end differentiable.
Our method can be used in conjunction with many networks; in particular, we study its application to the widely known MAC architecture, obtaining a significant reduction in the number of recurrent steps needed to achieve similar accuracies, therefore improving its performance to computation ratio.
Furthermore, we show that by increasing the maximum number of steps used, we surpass the accuracy of even our best non-adaptive MAC in the CLEVR dataset, demonstrating that our approach is able to control the number of steps without significant loss of performance.
Additional advantages provided by our approach include considerably improving interpretability by discarding useless steps and providing more insights into the underlying reasoning process.
Finally, we present adaptive computation as an equivalent to an ensemble of models, similar to a mixture of expert formulation.
Both the code and the configuration files for our experiments are made available to support further research in this area \protect\footnotemark.
\footnotetext{https://github.com/ceyzaguirre4/DACT-MAC}

\end{abstract}

\section{Introduction}

In the past few years, deep learning (DL) techniques have achieved state-of-the-art performance in
most, if not all, computer vision tasks \cite{AlexNet, Yolo, MaskRCNN, CompositionalAttention}.
While these methods are powerful in terms of representational capacity, they lack a suitable mechanism to allocate computational resources according to the complexity of each particular inference.
In effect, most popular DL based models used in computer vision applications, such as CNN
\cite{DeepLearning:Science:2015}, RNN \cite{DeepLearning:Science:2015}, Transformer
\cite{Transformer:2017}, and MAC \cite{CompositionalAttention} have a fixed processing pipeline
whose depth is independent of the complexity of the current input/output relation.

\begin{figure}[t]
 \begin{center}
 \includegraphics[width=1.0\linewidth]{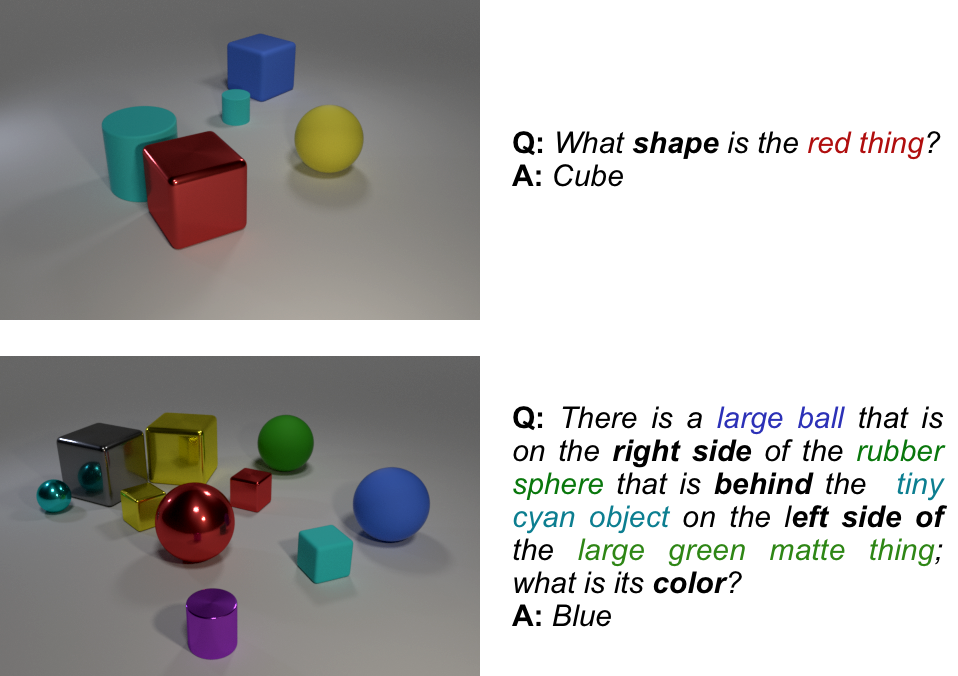}
 \end{center}
 \caption{Examples of questions in the CLEVR \cite{clevr} dataset that show a significant variation
in the number of reasoning steps that are needed to answer them correctly.}
\label{fig:clevrEg}
\end{figure}

The drawbacks associated with the use of a fixed processing pipeline can be illustrated by considering
tasks that require a complex sequential inference. This is the case of new Visual Question
Answering (VQA) scenarios that have been recently proposed to support research in the area of
visual reasoning, such as the CLEVR and GQA datasets \cite{clevr, gqa}.
These datasets pose challenging natural language questions about images whose solution requires the use of perceptual
abilities, such as recognizing objects or attributes, identifying spatial relations, or implementing
high-level capabilities like counting. As an example, Figure \ref{fig:clevrEg} shows two instances
from the CLEVR dataset \cite{clevr}. In this case, each visual question entails a different level
of complexity to discover the correct answer. Specifically, while the first question involves just
the identification of a specific attribute from a specific object, the second question requires the
identification and comparative analysis of several attributes from several objects. Despite this
significant difference, current visual reasoning models use the same processing pipeline to answer both questions.

From the previous example, it is possible to foresee that computational
efficiency, at training and inference time, is a relevant disadvantage of using a fixed processing pipeline. In effect, the usual goal of a DL model is to maximize accuracy, and as a consequence, the model is forced to calibrate its processing structure according to the most complex cases, overestimating the computational load needed to solve easier ones. This lack of computational efficiency not only causes longer processing times, but it also has major implications in terms of the environmental impact of AI
technologies, a problem that is gaining considerable attention \cite{strubell2019energy, Schwartz:EtAl:GreenAI:2019}. As an example, \cite{strubell2019energy} provides an estimation
of the carbon footprint of several NLP models, concluding that current AI models are becoming
environmentally unfriendly. This stresses the need to provide current DL models with a suitable
adaptive mechanism to control the computational effort needed to generate each
inference.
Furthermore, besides computational efficiency, as we show in this work, the use of an
adaptive processing pipeline might also play an important role in improving the overall accuracy of a
model and to improve its interpretability.

The previous argumentation highlights the need for suitable mechanisms to control computational
complexity in DL models; however, so far, research in this area has been limited. Soft attention
\cite{Bahdanau:MemNWs:2015} and skip connection mechanisms \cite{he15deepresidual} appear as
possible options to improve the efficiency of current DL architectures. These strategies, however, do
not allow to save computation, as they still require executing the full processing pipelines to
select attention areas or skip connections. Modular approaches are also an option
\cite{Andreas_2016,iep}. In this case, from a collection of specialized processing modules, a
controller or program generator adaptively selects on the fly a suitable configuration to handle
each query. Unfortunately, this strategy does not scale appropriately with the number and diversity of
modules needed to solve a task, which are usually limited to a fixed predefined collection. As an
alternative, instead of using specialized modules, recent approaches use a general-purpose neural
module that is applied sequentially to the input \cite{CompositionalAttention, NSM}. In this case,
each step in the sequence is expected to execute an operation necessary to arrive at the correct answer.
The specification of the number of steps needed to answer each question, however, is
non-trivial, so these models usually fix this value using a hyper-parameter for the whole model.

In a seminal paper \cite{DBLP:journals/corr/Graves16}, Graves introduces Adaptive Computation Time, ACT, an algorithm to adaptively
control computational complexity for Recurrent Neural Networks (RNN). The key idea behind ACT is to add a sigmoidal halting
unit to an RNN that, at each inference step, determines if the RNN should stop
or continue its processing. As an extra role, the activation values of the halting
unit are also used to ensemble the final output of the model as a weighted sum of
intermediate states. \cite{DBLP:journals/corr/Graves16} demonstrates the advantages of ACT by
showing its impact on improving the computational efficiency of RNN models on several synthetic
tasks. Posterior works have also shown the advantages of ACT when it is applied in real-world
scenarios, in the context of language modeling \cite{UniversalTransofmers, adaptiveReasoning} and
image understanding \cite{adaptiveReasoning}. Despite this success, as we show in this
work, ACT suffers from instability problems to find a suitable number
of steps to stop processing. We believe that this is due to its non-differentiable nature as ACT
achieves halting by forcing the sum of the weights used to ensemble the final output be equal
one by using a non-differentiable piecewise function.

Motivated by the evident advantages that a mechanism such as ACT might provide
to modern module networks, we propose a new approach to adaptive computation based on a novel
attention-based formulation. As key insights, this mechanism addresses two main problems of
ACT: i) Its non-differentiability and ii) The assumption that the hidden states of recurrent
architectures can be linearly combined. Our approach overcomes the non-differentiable operation in
ACT by using a halting unit to calculate, at each step, a running approximation of the final
output of the model. This approximation leads to a monotonically decreasing probability
distribution over all the outputs that implicitly includes a residual estimation of the
benefits of continuing the processing steps. This allows us to calculate an upper
bound for the future gain provided by subsequent outputs of the model. The
result is a fully differentiable model that can be trained using gradient descent whose
computation can be reduced by mathematically determining when an interruption of the processing
pipeline does not harm its final performance.
Finally, in our formulation, we also formalize adaptive computation as a
\textit{gate-controlled bucket of
models} that resembles the operation of a \textit{mixture of
experts} model \cite{MoE}.

As a testbed for our formulation, we choose the MAC network \cite{CompositionalAttention}, an
architecture that is gaining considerable attention for solving tasks that require visual reasoning. We highlight, however, that our mathematical formulation is highly general, and that it can also be applied to other learning architectures.
Our main results indicate that using the proposed method for adaptive computing achieves better performance than the same architecture with a comparable fixed computational cost.
Remarkably, the resulting models learn subjacent patterns in the data, as shown by the strong correlation between the number of steps executed on inferences and the template used to generate it, correlations which we then exploit to improve the interpretability of the model.
All this stands in stark contrast to the results obtained from the use of ACT, which fails to improve results or even meaningfully adapt the number of steps.

In summary, the main contributions of this work are: (i) A novel formulation of an
adaptive computational mechanism that is fully differentiable and it can be incorporated to
current DL based models, such as MAC network; (ii) A comparative analysis of the performance of our proposed method and ACT, demonstrating the superior performance of the first; (iii) An extensive experimental
evaluation demonstrating that the use of our adaptive computational mechanism on top of the MAC
network can increase not only computational efficiency but also performance and interpretability
of the resulting model.

\section{Related work}
Recent works have pointed out the need to improve the computational efficiency of DL models
\cite{strubell2019energy,Schwartz:EtAl:GreenAI:2019}. As an example,
\cite{strubell2019energy} shows a surprising estimation related to the high carbon footprint
of current DL techniques. Similarly, \cite{Schwartz:EtAl:GreenAI:2019} argues about the relevance
of including computational efficiency as an evaluation criterion for research and applications
related to artificial intelligence. In spite of this increasing need, research to improve
computational efficiency of DL models is still limited.

In terms of deep convolutional models, there have been works that attempt to
control the depth of these models, however, the main focus has been on improving accuracy
but not computational efficiency. As an example, approaches such as skip connections
\cite{he15deepresidual}
still require to compute a full model. Compact CNN models have also been explored in the context of visual recognition \cite{Zhou2016,Wen2016,Changpinyo2017}. As an example, sparsity constraints have been used to control the total
number of active parameters in a network \cite{Changpinyo2017}. This is an effective strategy to reduce the
computational complexity of a model, however, it consists of a global constraint that does not
adapt dynamically to each input. Attention mechanisms appear as an attractive option to focus computation in relevant information areas of the input, however, current attention techniques,
such as soft-attention \cite{Bahdanau:MemNWs:2015} or self-attention \cite{Transformer:2017}, focus
also on performance, requiring to execute the full processing pipeline.

In the context of recurrent networks, Graves proposes ACT \cite{DBLP:journals/corr/Graves16}, an algorithm designed to provide a RNN with a mechanism to dynamically adapt computational
complexity. Specifically, ACT attempts to dynamical allocate the proper amount of
computation for each particular input sample. The main challenge is that the complexity of each
input is unknown before attempting to build a suitable output. ACT handles this issue by adding a halting unit whose activation determines if the RNN should stop or continue with another
processing step. These activation values are then used to construct the models final output, as a
weighted sum of intermediate states of all previous recurrent steps. This is performed through a
series of non-differentiable operations mainly used to enforce a hard limit so that no subsequent
iteration changes the model output. As we show in this work, this results in noisy gradients that do
not handle properly the information about the number of processing steps being used.

In spite of its limitations, ACT has been applied to multiple tasks beyond the synthetic cases
reported in the original work \cite{DBLP:journals/corr/Graves16}. It has been used to improve
results on the LAMBADA language modeling dataset using a Universal Transformer architecture \cite{UniversalTransofmers},
achieving a new state-of-the-art performance. Also, on the challenging task of character level
language modeling, it has been used to dynamically increase the attention span of a Transformer
model, achieving state-of-the-art performance on the text8 and enwiki8 datasets. Furthermore, on the natural language reasoning corpus SNLI dataset, it has
been reported to boost performance and interpretability \cite{adaptiveReasoning}. In terms of
visual recognition, \cite{adaptiveResNet} proposes a DL architecture based
on residual networks that uses ACT to dynamically choose the number of executed layers for
different pre-defined regions in the input image. \cite{adaptiveResNet} applies this strategy to
the case of visual classification, reporting improving performance in terms of computational
efficiency and
model interpretability.

Our approach to adaptive computation has substantial differences with respect to ACT. ACT achieves
halting by forcing that the weights used to combine each step's output into the final answer sum exactly one. To attain this behavior a
non-differentiable piecewise function is used, namely: if the sum of the weights is more than one, then change the last weight so that the sum is exactly one. In contrast, our approach
maintains the full gradient by only halting during evaluation (and not during training). The weights
used to combine all the step output's are described by a monotonically decreasing probability
distribution that implicitly includes future steps yet to be computed. The result is a fully
differentiable model for training with gradient descent whose computation can be reduced
during inference by mathematically determining when the interruption cannot change the output.

In terms of modular networks, existing approaches can be divided into those that combine
multiple specialized modules \cite{Andreas_2016, iep}, and those that use a single general purpose
module \cite{CompositionalAttention, NSM}. In the case of specialized modules, the generation of
the sequences required costly supervision or elaborate reinforcement learning training schemes. In the case of a general purpose module, the selection of the module to execute is trivial (only one), however, the number of steps to apply this module cannot be determined for each sample,
instead its value is fixed as a hyper-parameter. In this work, we build upon these networks by
replacing this fixed hyper-parameter by an adaptive approach to select the horizon of the
computational pipeline.

\begin{figure*}
  \begin{center}
      \includegraphics[width=0.9\linewidth]{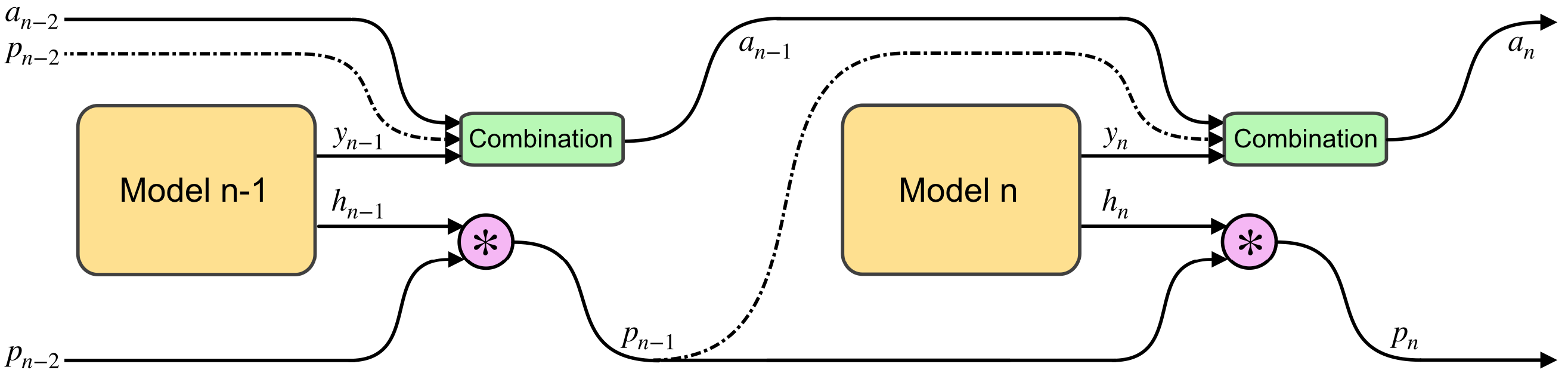}
  \end{center}
  \caption{
    The \textit{accumulated output} $a_n$ is built by linearly combining $a_{n-1}$ with the output of the $n$th model (following Equation \ref{eq:2}).
    Each step can limit the contribution of future steps by maintaining or reducing the value of the scalar $p_{n-1}$ used for the linear combination (illustrated with a dotted line).
    Any $h_n$ valued roughly zero will force $p_n$ to this value, effectively disallowing the outputs of future models from altering the current \textit{accumulated output}, and effectively imposing that this $a_n$ become the final output $Y$.}
\label{fig:0}
\end{figure*}

\section{Differentiable Adaptive Computation Time}
\label{model}

In this section, we present our method to implement a Differentiable Adaptive Computation Time
strategy (DACT).
Our formulation can be applied to any model or ensemble that can be decomposed
as a series of modules or submodels $m_n$, $n \in [1,\dots,N]$ that can be ordered by complexity.
For example, recurrent networks are composed by iterative steps, CNNs by residual blocks, and ensembles by smaller models.
We refer to the composition as our final model or ensemble $M$, and to its output as $Y$.

In the context of VQA, $m_n$ receives as an input question $Q$ and image $I$, as well
as any relevant representation from a previous submodel $m_r$ , $r<n$. Each submodel $m_i$ should
produce its own prediction $y_n$ about the correct answer to $Q$. Additionally, each submodel
$m_n$ should also produce a sigmoidal output $h_n \in [0,1]$ that represents how uncertain is $m_n$
about the correctness of its output $y_n$, where we define the initial value $h_0=1$.

The use of scalars $h_n$ is the main mechanism to provide $M$ with an adaptive computation
mechanism. The key idea is to restrict models with higher index $s > n$ from
altering the final answer of $M$ once the current uncertainty about the correct answer is below a
target level. With this goal in mind, let define:

\begin{equation}
    \label{eq:1}
    p_n = \prod_{i=1}^{n}h_{i} = h_{n} p_{n-1}
\end{equation}

The value of $p_n$ can be interpreted as the probability that a subsequent submodel $m_s$, $s >
n$ might change the value of the final answer $Y$ of the ensemble $M$. Consequently, we
define the initial value $p_0=1$.

According to the previous formulation, $h_n$ represents the uncertainty of submodel
$m_n$, while $p_n$ represents the uncertainty of the full ensemble considering the first $n$ models. From
Eq. (1), it is easy to see that the values of $p_n$ are monotonically decreasing with respect
to index $n$. Also, notice that a small value of $h_n$ forces the future values of $p_n$ to be
close to $0$.

We still need to describe how to combine all \textit{intermediate outputs} $y_n$ ($n \in [1,\dots,N]$) to form $Y$.
We achieve this by defining auxiliary accumulator variables $a_n$ which contain the ensemble's answer up to
step $n$. By using Eq. \ref{eq:1}, we can construct $a_n$ in such a manner that for some step $n$
with a low associated $p_n$ then $a_n \approx Y$:
\begin{equation}
    \label{eq:2}
    a_n = \begin{cases}
    \overrightarrow{0} \qquad \text{if} \quad n=0\\
     \evy_n p_{n-1} + a_{n-1} \left( 1 - p_{n-1} \right) \quad \text{otherwise}
    \end{cases}
\end{equation}

It follows from this definition that $Y$ can always be rewritten as a weighted sum of
\textit{intermediate outputs} $y_n$.
Additionally, the sum of the weights is always equal to $1$, thus describing a valid
probability distribution over the \textit{intermediate outputs} $y_n$.
Both proofs are included in supplementary material.

Therefore, by describing what is effectively a \textit{pair-wise} linear interpolation, we obtain a
method for \textbf{implicitly attending} the outputs of each model in the ensemble, including
succeeding ones. In this manner, what we propose is essentially a \textit{mixture of experts} type
ensemble \cite{MoE} where we remove the controller and replace the gating model for the implicit distribution
described above. As a main result, by adding probabilities instead of hidden values as in ACT, we
remove the assumption of ACT that the hidden states of the underlying RNN are approximately linear.

No restriction is placed on whether two models in the ensemble can communicate so long as the origin of the exchange is always before in the ordered model sequence as is always the case with recurrent architectures.

\subsection{Penalizing complexity}
\label{penalizingcomplexity}

Following the principle of Ockham's razor, we wish to reduce complexity when it is not needed by choosing simpler models in lieu of more complex ones when both provide similar results.
To achieve this, we define the ponder cost $\rho$ as:
\begin{equation}
    \label{eq:3}
    \rho = \sum_{n=1}^{N} p_n
\end{equation}

By adding the \textit{ponder cost} to the loss function $L$ we encourage the network to minimize the contribution of more complex models.
This is used in the next section \ref{cactbehaviour} to reduce computation.
\begin{equation}
    \label{eq:4}
    \hat{L} (\vx, \vy) = L (\vx, \vy) + \tau \rho(\vx)
\end{equation}
where $\tau$ is the \textit{time penalty}, a hyper-parameter used to moderate the trade-off between complexity and error.

\subsection{Reducing computation time}
\label{cactbehaviour}

The previous formulation allows us to train a model incorporating the DACT methodology. In other
words, we modify the training process of the model to allow simpler models to cap the maximum
impact of all subsequent ones (equations \ref{eq:1} and \ref{eq:2}). As a consequence, we can
avoid running more complex models when they cannot significantly change the final output $Y$. In
this Section, we show that at test time we can use a halting criterion to ensure that the
subsequent steps of the sequence do not change the current prediction.

The choice of the criteria for halting (and therefore reducing computation) depends greatly on the
task and how close of an approximation is required. In this work, our goal is to achieve
the same top-1 accuracy with and without using DACT. This is equivalent to establish a
halting criterion such that the class with highest probability in $a_n$, i.e. using $n$ sub-models,
 will be the same as that in $Y$.

We know that $y_n$ (the \textit{intermediate output} of the $n$th classification model) is
restricted to  $0 \leq y_n \leq 1$ as a result of using either \textit{Softmax} or \textit{Sigmoid}
functions. Since the maximum change of the accumulated answer $a_n$ in the
remaining $d = N - n$ iterations is limited by $p_{n}$, we can calculate the maximum difference
between the predicted probabilities for the top most class and the \textit{runner-up}.
Consequently, we can achieve reduced computation by halting once this difference is
insurmountable.

Without loss of generality consider the case where, for some step $n$, the class
with the highest probability in the accumulated answer $a_n$ corresponds to class $c^*$ with
probability $\Pr(c^*, n)$, and the runner-up (second best) class is $c^{ru}$ with probability
$\Pr(c^{ru}, n)$. The minimum value for the probability of the class $c^*$ after the $d$
remaining steps is obtained when all the future steps assign a
minimum probability ($0$) to this class. We can use this result to obtain a lower bound to the probability:
\begin{equation}
    \label{eq:6}
    \Pr(c^*, N) \geq \Pr(c^*, n) \prod_{i=n}^{n+d-1}(1-p_i)
\end{equation}

Leveraging that $p_{n} \geq p_{n'}$ (for any $n'$ greater than $n$) in conjunction with
Eq. \ref{eq:2}, we can establish that the minimum value for the class at $c^*$ after another $d$ steps
is always:
\begin{equation}
    \label{eq:6}
    \Pr(c^*, N) \geq \Pr(c^*, n)(1-p_n)^d
\end{equation}

Likewise, the maximum value that the probability for the runner-up class $c^{ru}$
can take after all unused $d$ steps ($\Pr(c^{ru}, N)$) is achieved when the maximum
probability ($1$) has been assigned to this class at every remaining step.
Replacing this value into Eq. \ref{eq:2} yields an upper bound to the value that the probability
for the class $c^{ru}$ can take:
\begin{multline}
    \label{eq:8}
    \Pr(c^{ru}, N) \leq \Pr(c^{ru}, n) \prod_{i=n}^{n+d-1}(1-p_i) \\ + \sum_{i=n}^{n+d-1}
p_i \prod_{j=i+1}^{n+d-1}(1-p_j)
\end{multline}

Then, since $0 \leq p_n \leq 1$ and  $p_{n} \geq p_{n'}$ ($\forall n' \ge n$), we obtain that the maximum value for the class $c^{ru}$ is:
\begin{equation}
    \label{eq:8}
    \Pr(c^{ru}, N) \leq \Pr(c^{ru}, n) + p_n d
\end{equation}

We say that the difference between the top class and the runner up is insurmountable once we prove
that $\Pr(c^*, N) \geq \Pr(c^{ru}, N)$, and thus we can cut computation
since the remaining steps cannot change the final answer of the model.
Mathematically, this means the \textit{halting condition} is achieved when:
\begin{equation}
    \label{eq:9}
    \Pr(c^*, n)(1-p_n)^d \geq \Pr(c^{ru}, n) + p_n d
\end{equation}
which is the criterion used in this work to stop processing.

\begin{figure*}
\begin{center}
  \includegraphics[width=0.9\linewidth]{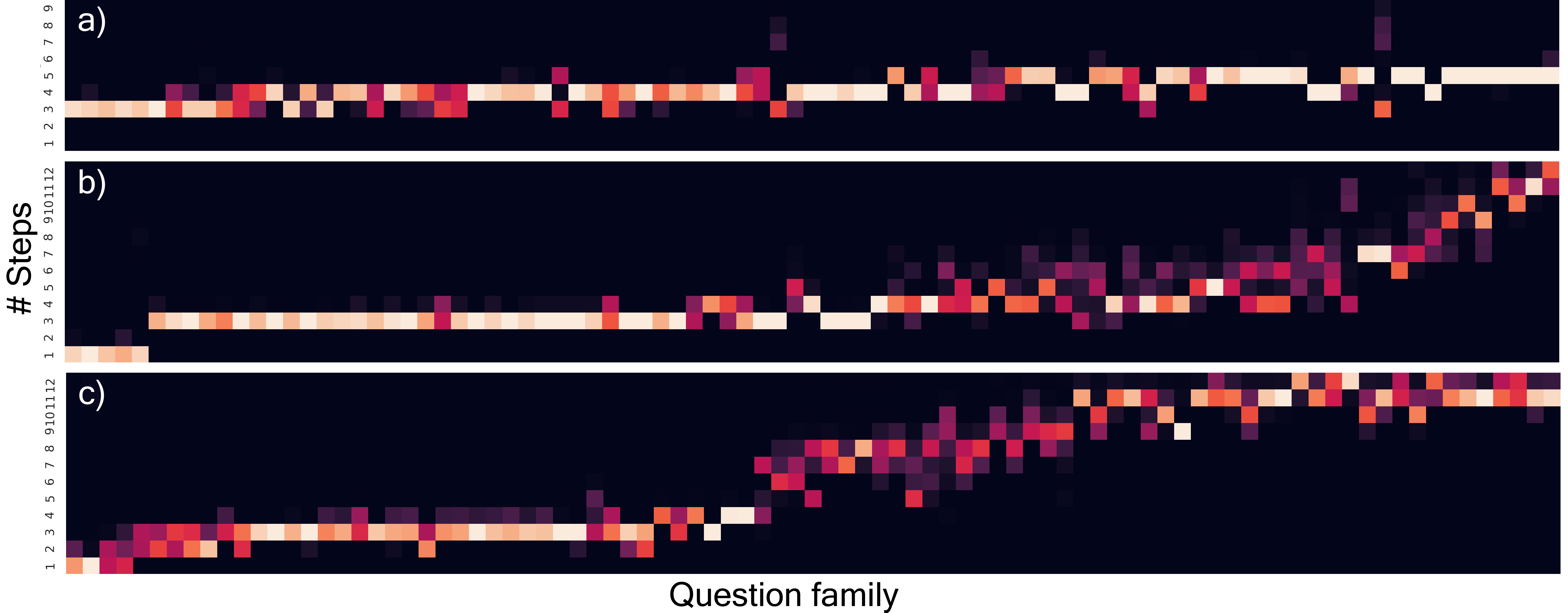}
\end{center}
\caption{Questions in CLEVR are synthetically generated following templates, for example, by replacing $<$C$>$ and $<$M$>$ with a color and material in the template \textit{“How many $<$C$>$ $<$M$>$ things are there?”}.
Accordingly, adding adaptability to the model does not increase performance but rather, similar complexity to solve.
The figure shows the average amount of computation used by three models for each question family, sorted by the average number of steps used by the respective model. The first image (a) illustrates how ACT fails to learn how to answer the most straightforward questions in less than three steps, or the hardest in more than five \protect\footnotemark.
Below it, b) shows the results for a variant of DACT that averages approximately the same number of steps but uses more of the available spectrum, significantly improving model performance.
The last image shows a variant of DACT, which uses 50\% more reasoning steps on average and thus achieves even better performance.
}
\label{fig:corr_compare}
\end{figure*}
\footnotetext{This ACT variant was cherry-picked as it achieved the highest accuracy while also doing the maximum amount of steps observed for ACT.}

\begin{figure}[t]
\begin{center}
  \includegraphics[width=1.0\linewidth]{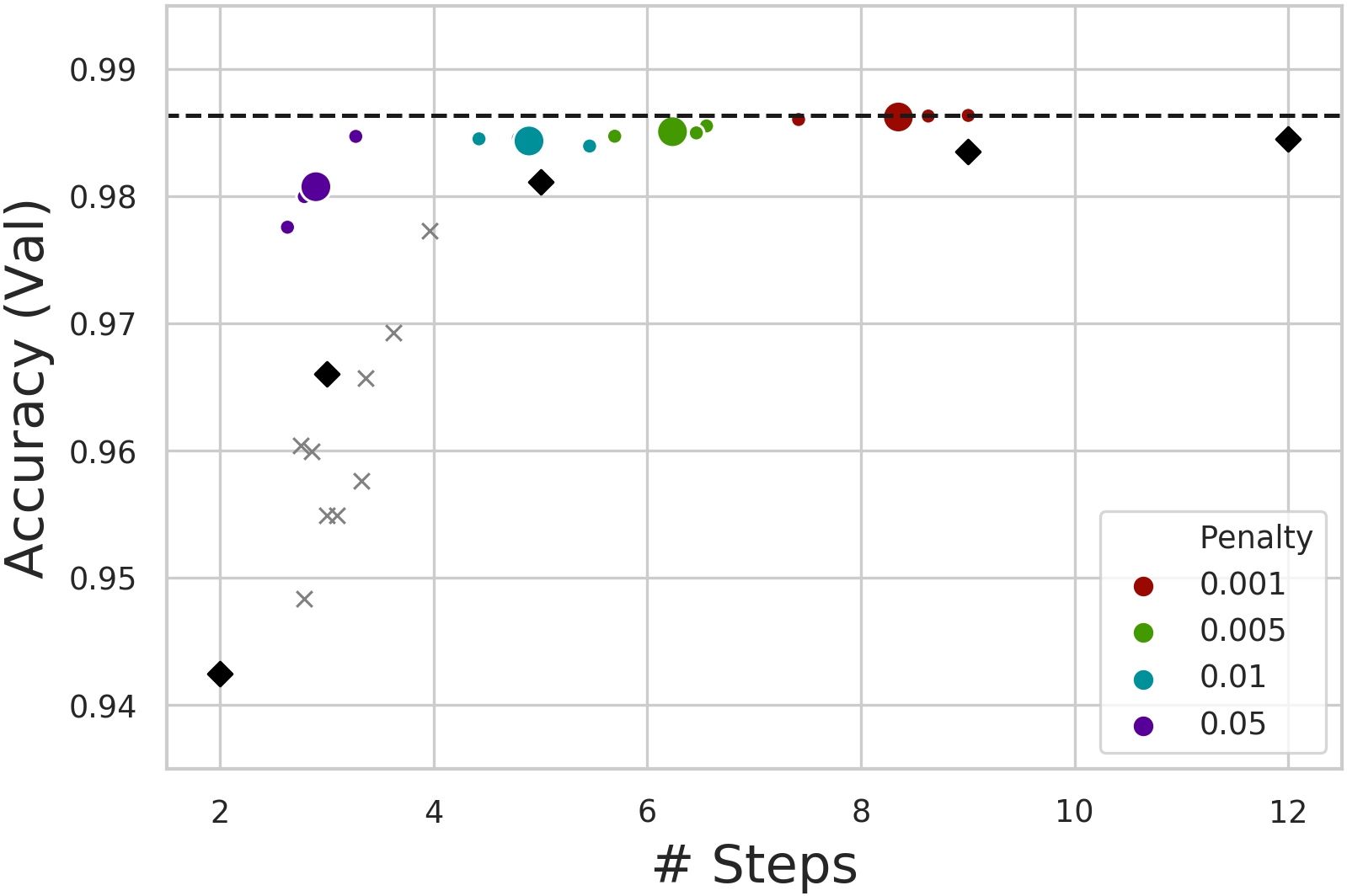}
\end{center}
\caption{Scatterplot showing the relationships between computation (measured in average steps, horizontal), and precision (measured in accuracy, vertical)for each model, where every experiment was repeated three times.
The results obtained with DACT are shown in color, with individual runs represented as small circles while the averages for each penalty are shown as larger ones.
The averaged results for ACT are shown as gray \textit{X}es. No color is used as the value for the ponder cost did not impact the number of steps.
The diamonds show the average accuracy obtained by MAC at different network lengths, while the dotted line represents the accuracy of the best performing 12 step MAC.}
\label{fig:mac_steps_acc}
\end{figure}

\begin{figure}[t]
\begin{center}
  \includegraphics[width=1.0\linewidth]{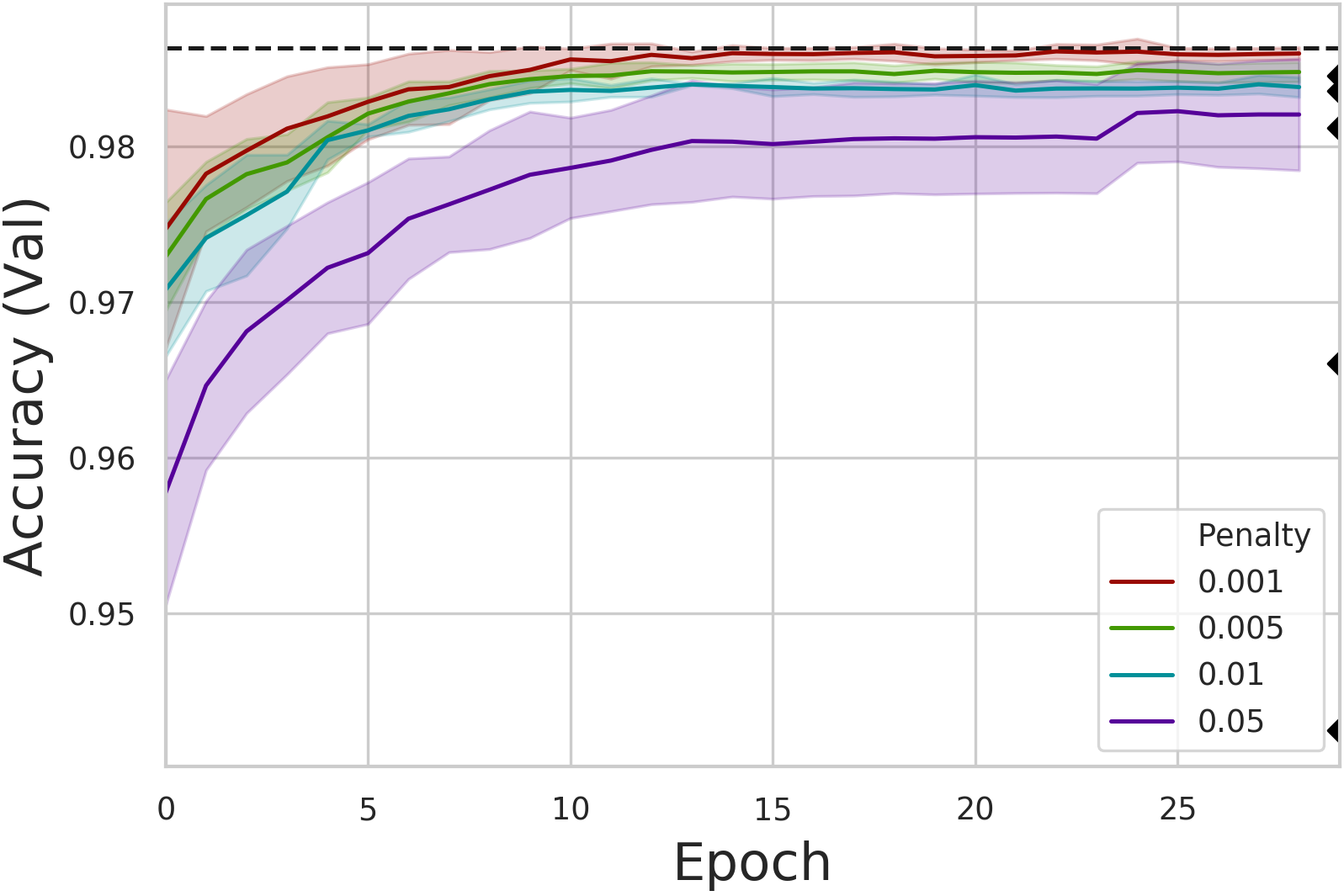}
\end{center}
\caption{Learning curves of DACT enabled MACs with different ponder costs.
For reference, we include the maximum accuracy achieved by any non-adaptive MAC as a dotted line. The black marks show the average accuracy obtained by this model when restricted to 12, 9, 5, 3, and 2 steps from top to bottom, respectively.
Recall that these models are first pre-trained for ten epochs without any gating or adaptive algorithms.}
\label{fig:convergence}
\end{figure}

\section{Experiments}
\label{exp:vqa}

The MAC network is a state-of-the-art recurrent architecture that decomposes problems into reasoning steps.
This network iterates for a fixed number of times (usually 12) where each step first attends the question, then the image, and finally, it updates an internal memory representation.
When applied to the CLEVR \cite{clevr} dataset, the MAC sets state-of-the-art performance with 98.9\% accuracy.

We start from a PyTorch \cite{pytorch} port of MAC available online \footnote{https://github.com/ceyzaguirre4/mac-network-pytorch}, which we trained without self-attention to achieve a maximum accuracy score of 98.6\% on CLEVR overall.
To help convergence and speed up training, we first pre-train a variant of the model on CLEVR without gating or self-attention for ten epochs (with all hyper-parameters set to their defaults).
We then reset all optimizers and train three main variants starting from the saved weights.
First, we add the gate to the MAC, slightly improving the results.
Second, we train several ACT versions using different ponder costs.
Finally, we do the same with DACT.
All variants are trained for another additional 30 epochs, saving the weights with the highest associated accuracy on the validation set.

As one of the main goals of adaptive computation is to maximize performance at lower computational cost, we evaluate each model's accuracy with respect to the average number of steps taken to reach the best score.
As illustrated in Figure \ref{fig:mac_steps_acc}, models resulting from the application of DACT to MAC substantially outperform non-adaptive versions of MAC with similar computation cost in the CLEVR dataset (achieved by training a MAC with the number of steps fixed to the closest integer).
Additionally, in our experiments, DACT trained with a ponder cost of $\lambda = \num{1e-3}$ repeatedly obtains an accuracy comparable to the best achieved by any MAC and, on average, surpasses all tested alternatives.
This apparent contradiction (obtaining better results with less computation) can be explained by considering that DACT-augmented-MACs have the same representational capacity as regular MACs, but can choose to reduce computation when needed.

The same results also show that, when provided with sufficient resources, MAC increases its performance reducing its gap with respect to DACT versions. This tendency, however, does not hold beyond 12 iterations, as shown also in \cite{CompositionalAttention}.
We train a 15 step MAC with gating using the same training scheme, and the results are worse than those from its 12 step counterpart, revealing the inadequacy of the gating mechanism.
In contrast, DACT-enabled-MACs with the maximum amount of steps set to 15 can be fine-tuned from existing 12 step models to obtain the best results of any model tested at 98.72\% accuracy.
In addition to improving performance, these results prove that using our algorithm on MACs makes them more robust to increase the value of the maximum number of steps.

On the other hand, models trained with the existing algorithm (ACT) are unsuccessful in surpassing the accuracy of computationally equivalent MACs. In particular, DACT responds as expected to variations in the ponder cost, adapting its computation accordingly, however, ACT proves to be insensitive to the ponder cost. As an example, a variant of ACT without ponder cost ($\lambda = 0.0$) performs $3.2$ steps on average and obtains an accuracy of $95.8\%$.

We also evaluate how well the model adapts to variations in question complexity, since the rationale behind adapting the number of steps is to enable the models to allocate more computation to more complex questions. As expected, DACT iterates fewer times for easy questions and more when the input question is more complex, improving model performance at no additional computational cost.
In Figure \ref{fig:corr_compare}, questions are clustered by family type which translates to groups that require similar step sequences to solve and therefore are of similar complexity (the figure is further explained in supplementary material, where we include examples for each family).
This figure shows a remarkable correlation between computation and question complexity, despite not including any type of supervision about these factors.

Finally, in order to evaluate the generality of the suggested approach to real data, we evaluate the combined DACT-MAC architecture on the more diverse images and questions in the GQA dataset \cite{gqa}.
We start by again pre-training a non-gated MAC (4 steps, 5 epochs) and then fine-tuning ACT, DACT and gated MAC variants for another 15 epochs.
The results shown in Table 1
of the supplementary material show that DACT is effective in reducing the number of steps needed while maintaining most of the performance of the architecture that always iterates the maximum number of times (four steps).
However, we found in our experiments that for GQA the chosen architecture (MAC) doesn't benefit from iterating more than two steps, and even then the advantage gained over its non recurrent single-step version is marginal.
Accordingly, adding adaptability to the model does not increase accuracy but rather results in a small but measurable reduction in performance.

Regardless of the above, the experimental results highlight the advantages of our algorithm with respect to ACT, showing that DACT once again obtains better results for the same number of steps.
Additionally, while our method continues to adapt computation in a coherent manner to the \textit{time penalties}, ACT remains mostly irresponsive to the values these take.
Furthermore, the high correlations between computation and question type are also present for the GQA dataset as Figure 2
of the supplementary material shows, revealing once more that DACT learnt to meaningfully adapt complexity without supervision.

\begin{figure}[t]
  \begin{center}
  \includegraphics[width=0.8\linewidth]{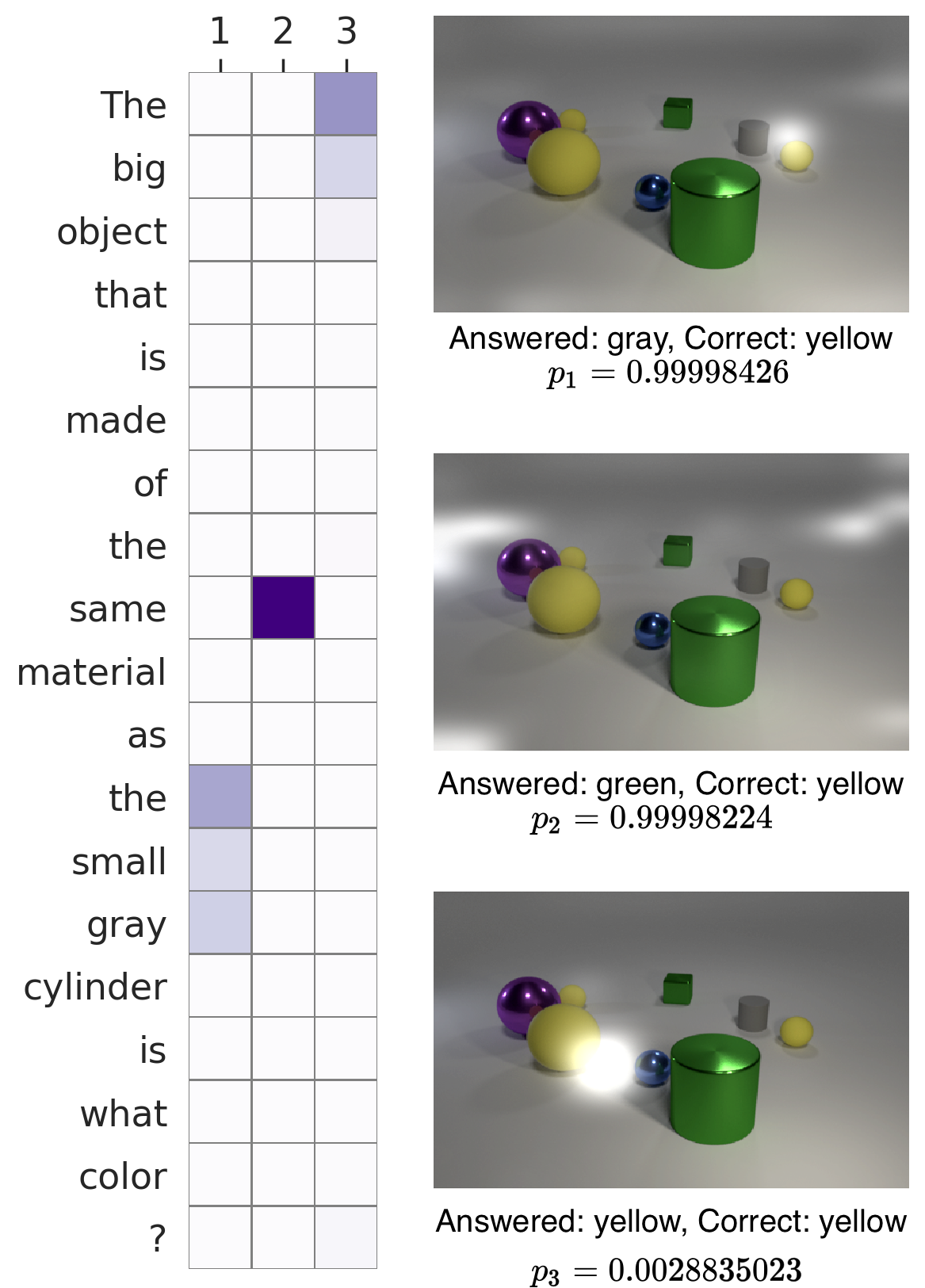}
  \end{center}
  \caption{Attention maps, intermediate answers, and halting probabilities captured from DACT for the image and question shown.
   Three steps were needed to arrive at the answer.
   The first two steps output wrong answers with high uncertainty ($p_n \approx 1$).
   The last step, however, has identified the relevant object and can thus answer correctly and with confidence.}
  \label{fig:MACAtts}
 \end{figure}

\section{Discussion}

\begin{figure*}
  \begin{center}
  \includegraphics[width=0.8\linewidth]{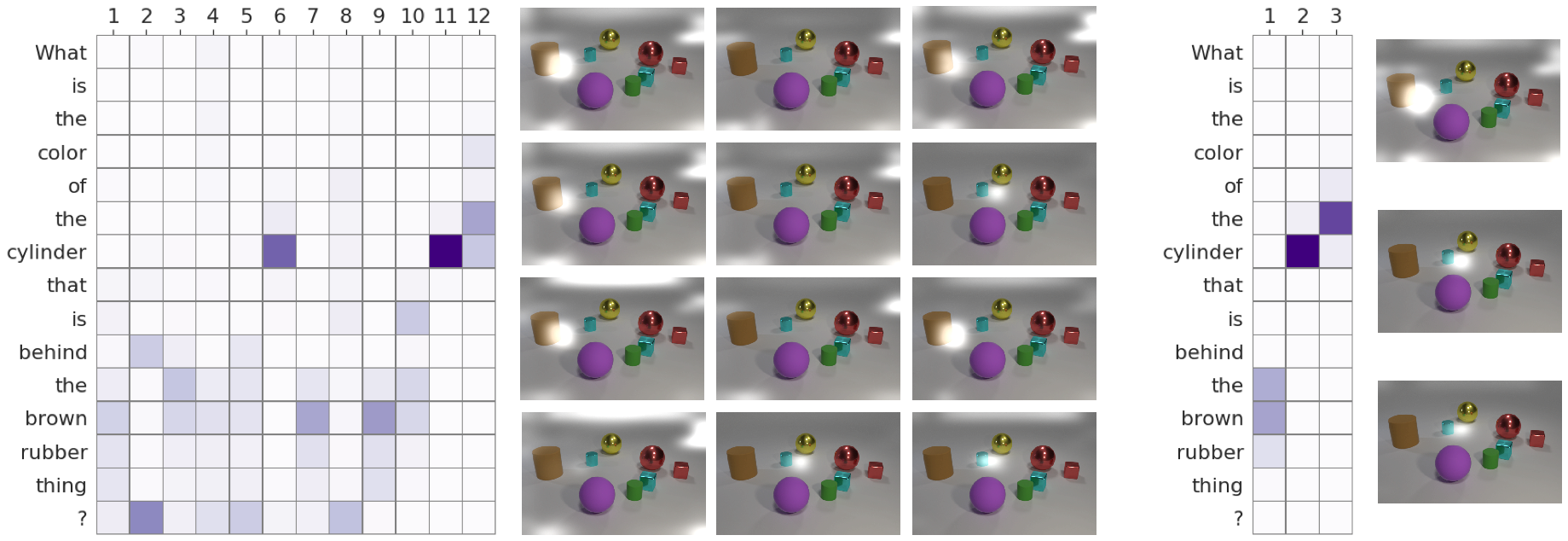}
  \end{center}
  \caption{Linguistic and visual attention maps for both the standard MAC architecture (left) and our DACT enabled variant trained with $\tau = \num{5e-3}$ (right).
  Besides the obvious and substantial reduction in the number of steps used to answer, our model also contributes to the overall interpretability of the inference.
  This is achieved by adding a proxy of the number of steps taken to the loss function, effectively coercing the model into only using fewer (and therefore more likely to be semantically strong) steps.
  The question attentions above show that the last two steps are similar for both models, but that only one of the other ten steps used by MAC was necessary.
  }
  \label{fig:AttsComp}
  \end{figure*}

As in previous works \cite{iep, CompositionalAttention}, we also analyze the attention maps provided by the model.
In particular, we examine both the linguistic and visual attentions generated at each step. Also, we raise the question of whether the proposed architecture can indeed improve interpretability.
Figure \ref{fig:AttsComp} shows examples of the attention maps generated by the 12 step MAC.
Since the MAC architecture only considers the last state in memory for the final classification, the final controls tend to be significant.
Indeed, our test indicates that the last few execution steps generate similar attention maps to those produced by our adaptive variant.
However, as Figure \ref{fig:corr_compare} shows, very few queries need all 12 steps of computation, so most of the steps execute either, repetitions of other operations, or are just padding (\eg attending punctuation).

The above stands in contrast to our DACT enabled variant, which in practice provides a \textit{free lunch} by maintaining the performance while increasing interpretability without (in the case of MAC) adding additional parameters.
We achieve this by adding the differentiable approximation to the number of steps taken, the \textit{ponder cost} (Eq. \ref{eq:3}), to the loss function.
Consequently, since the model is coerced into only using significant steps, we find that those taken are more likely to be semantically meaningful.

In addition, the formation of the final output of the model from the sub-outputs enables us to check what the model would answer at each timestep.
When analyzed in conjunction with the halting probabilities both yield valuable insights on the internal representation of the model. For instance, in Figure \ref{fig:MACAtts}, the first step has minimal information from the question and image and consequently is very uncertain of the given answer. However, this limited information is enough for the model to identify that the question involves the color of some object, and therefore the answer is the color of the only object it has seen.
We expect the increased transparency of the model will assist future studies on explainability and the detection of biases in datasets.

The difference in steps for distinct queries allows us to obtain an estimation of the complexity of each encoded question as shown in \ref{fig:corr_compare}.
These can then be combined to obtain an estimator for subsets of the dataset.
Estimators such as the average complexities of questions can be useful for curriculum learning \cite{curriculum_learning}, providing a novel way of separating data that does not require supervision.

Finally, due to its non-differentiable inclusion of the number of steps used, ACT effectively provides the same penalty to each step.
On the other hand, DACT is not limited in this respect, since probabilities $p_n$ are entirely differentiable.
This allows for the inclusion of functions of the amount of steps in the loss, opening another line for future work.
We expect that the inclusion of non-linear functions of the ponder cost (such as the square of its value) can also be a interesting avenue for future research.

\section{Conclusion}

This paper introduces a new algorithm for adaptive computation called DACT, which, unlike to existing ones, is end-to-end differentiable.
By combining our method with the MAC architecture, we manage to significantly improve its performance to computation ratio by reducing the number of recurrent steps needed to achieve a certain accuracy threshold.
Furthermore, we show that the inclusion of DACT improves the robustness of the resulting model to an increase in the number of processing steps,
improving performance with respect to previous state-of-the-art results in CLEVR.
Additionally, our results also show that DACT improves interpretability by providing additional insights into the internal operation of the MAC architecture, showing how its prediction and uncertainty change at the different steps of the estimation. As future work, we believe that our formulation of adaptive computation as an ensemble of models can motivate further research in this area.


\textbf{Acknowledgments:}
This work was partially funded by FONDECYT grant 1181739 and the Millennium Institute for Foundational Research on Data (IMFD).

\section{Supplementary Material}

\subsection{Implementation details}

Unless otherwise stated, all models use the default parameters from the official implementation of MAC \footnote{https://github.com/stanfordnlp/mac-network}.
Notable differences between these and those in the original paper \cite{CompositionalAttention} include using different dropout rates for different sections of the network (92\% on output of biLSTM, 82\% on convolutional filters), initializing weights by sampling from the Xavier uniform distribution \cite{xavier}, setting the question as the first control, and using a LR scheduler.
We use a gate bias of $1.0$ for the gated MAC as this achieves the best performance in the original paper; however, unless stated otherwise, we do not bias our adaptive computation algorithms.

\begin{figure}[t]
\begin{center}
  \includegraphics[width=1.0\linewidth]{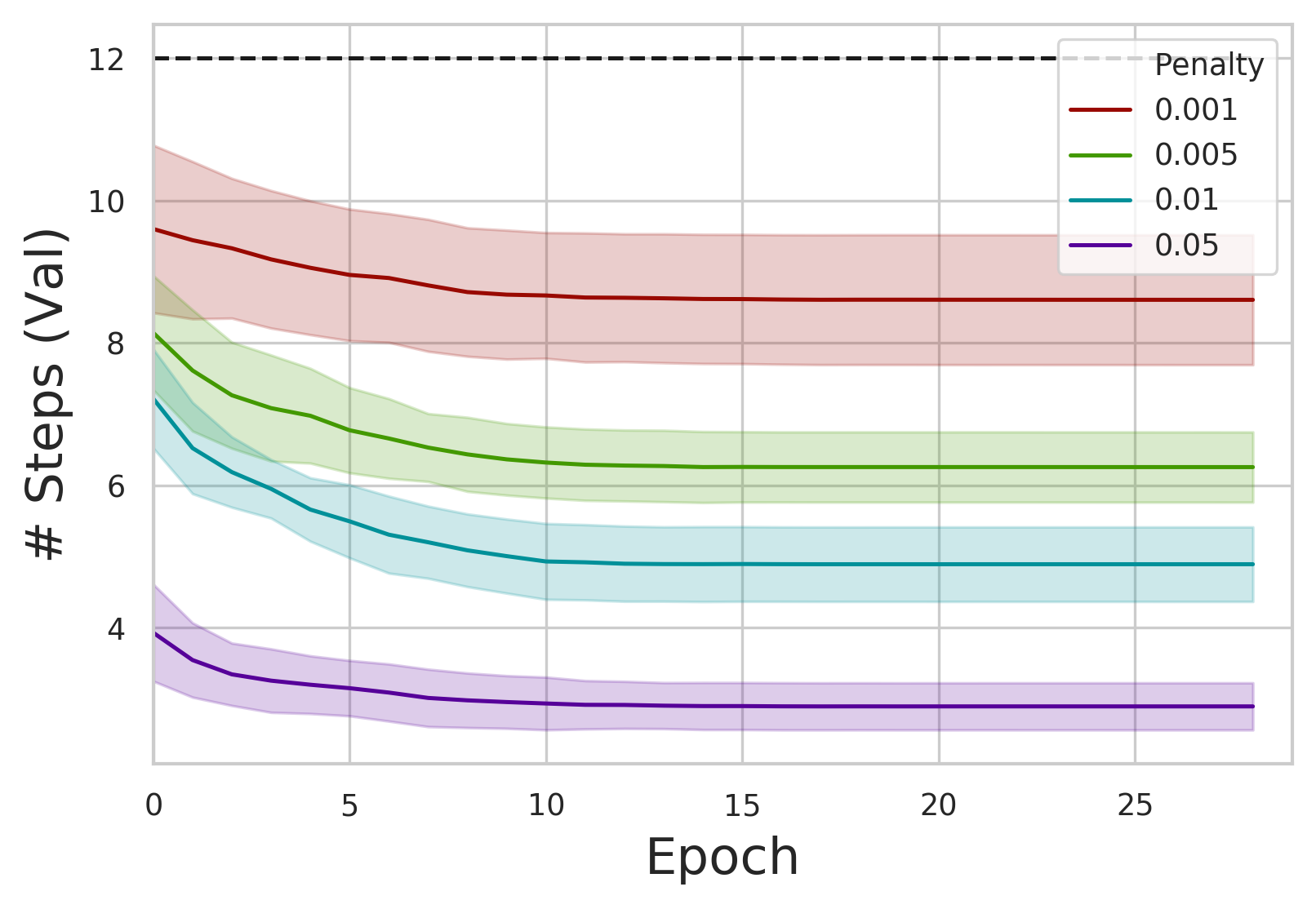}
\end{center}
\caption{Curves illustrate the change in the number of steps employed by DACT-MACs as fine-tuning on CLEVR progresses.
For reference, we include the case of standard non-adaptive MAC as a dotted line (12 steps).
Note that in the first epoch the model uses more steps than the value to which it converges, following a behavior similar to the pre-trained MAC, while subsequent epochs reduce computation.
Using higher ponder costs translates into less variability in the number of steps used, shown as translucent bands of the same color.
}
\label{fig:convergence_steps}
\end{figure}

Experimental results in Section 4 are obtained by training from scratch the MAC network three times and then using the resulting weights to train multiple other models.
Each of the pre-trained MACs are further trained to obtain: four variants of DACT, corresponding to different values of $\lambda$; and seven variants of ACT, six with different $\lambda$'s and one without ponder penalty.
Training the variants is done by reseting all schedulers and optimizers and then reinitializing a new MAC with its corresponding gating mechanism.
We then overwrite the pertinent weights with the pre-trained ones and train for another 30 epochs saving the best model according to the monitored precision.
This is done to reduce the time needed to train all models while still providing significant mean and variance values for relevant metrics (\ie precision and steps).
Figure \ref{fig:convergence_steps} shows  how the average number of steps used by DACT-MACs for CLEVR \cite{clevr} adapts from the original 12 step algorithm learnt during the pre-training phase of the 12-step non-adaptive MACs, progressively reducing the total computation needed by iterating fewer times.

Few changes were needed in order to apply adaptive MACs to GQA \cite{gqa}. Instead of using regions of the image (top-down attention), our knowledge-base is composed of region detections (bottom-up) \cite{anderson2018bottom}. We also use 1x1 convolution filters instead of 3x3 for the same reason.
Additionally, we use a 4 step non-adaptive MAC during the pre-training phase as this is the recommended number of steps for the model \cite{gqa} and also yielded the best results in our experiments.
The resulting accuracies, along with the mean number of steps used to attain them, are shown in Table \ref{table:mac_steps_acc_GQA}.

\begin{figure*}
  \begin{center}
      \includegraphics[width=0.84\linewidth]{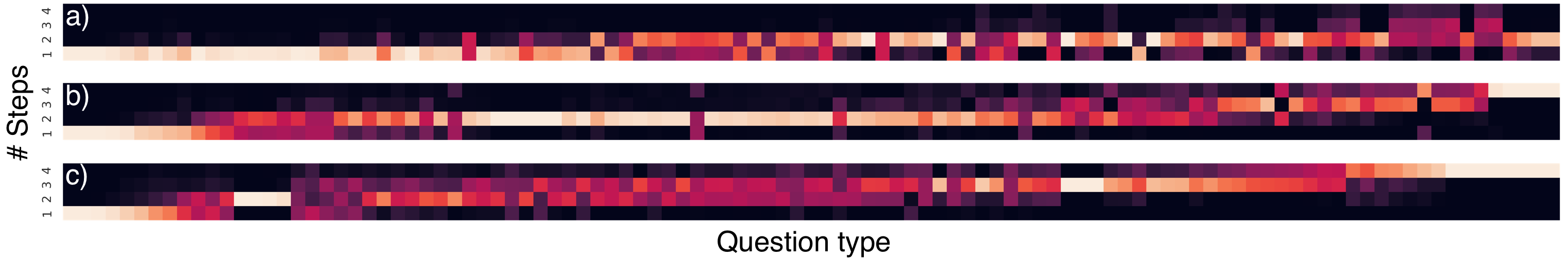}
  \end{center}
  \caption{
    The figure shows the the distribution of the number of steps used by DACT for each one of the 105 different \textit{question types} in the GQA dataset.
    In order from top (a) to bottom (c) we show how decreasing the \textit{time penalty} (\num{5e-2}, \num{1e-2}, \num{5e-3} for a,b, c respectively) results in increased total computation.
    }
\label{fig:corr_gqa}
\end{figure*}

Each question in GQA has been assigned one of 105 \textit{question types} according to the operations needed to answer the question.
Therefore, as was the case with question families in CLEVR, we group questions by type with the underlying assumption being that all questions of a type have similar complexities.
Figure \ref{fig:corr_gqa} validates that, as in CLEVR, DACT is adapting computation, and the total amount of computation varies following the time penalty.

\begin{table}
\begin{center}
\begin{tabular}{|l|c|c|c|}
\hline
Method & Ponder Cost & Steps & Accuracy \\
\hline\hline
MAC+Gate & NA & 2 & 77.51 \\
MAC+Gate & NA & 3 & 77.52 \\
MAC+Gate & NA & 4 & 77.52 \\
MAC+Gate & NA & 5 & 77.36 \\
\hline\hline
ACT & \num{1e-2} & 1.99 & 77.17 \\
ACT & \num{1e-3} & 2.26 & 77.04 \\
ACT & \num{1e-4} & 2.31 & 77.21 \\
ACT & 0 & 2.15 & 77.20 \\
\hline\hline
DACT & \num{5e-2} & 1.63 & 77.23 \\
DACT & \num{1e-2} & 2.77 & 77.26 \\
DACT & \num{5e-3} & 3.05 & 77.35 \\
DACT & \num{1e-3} & 3.69 & 77.31 \\
\hline
\end{tabular}
\end{center}
\caption{Our proposed method (DACT) achieved better accuracy than existing adaptive algorithms on the GQA \textit{test-dev} set, while also adapting computation coherently to the values taken by the ponder cost hyper-parameter.
However, the task did not benefit from increased computation, so all adaptive models incur in a small metric loss compared to non-adaptive variants.}
\label{table:mac_steps_acc_GQA}
\end{table}

\subsection{Question Families}

\begin{figure*}
  \begin{center}
      \includegraphics[width=0.84\linewidth]{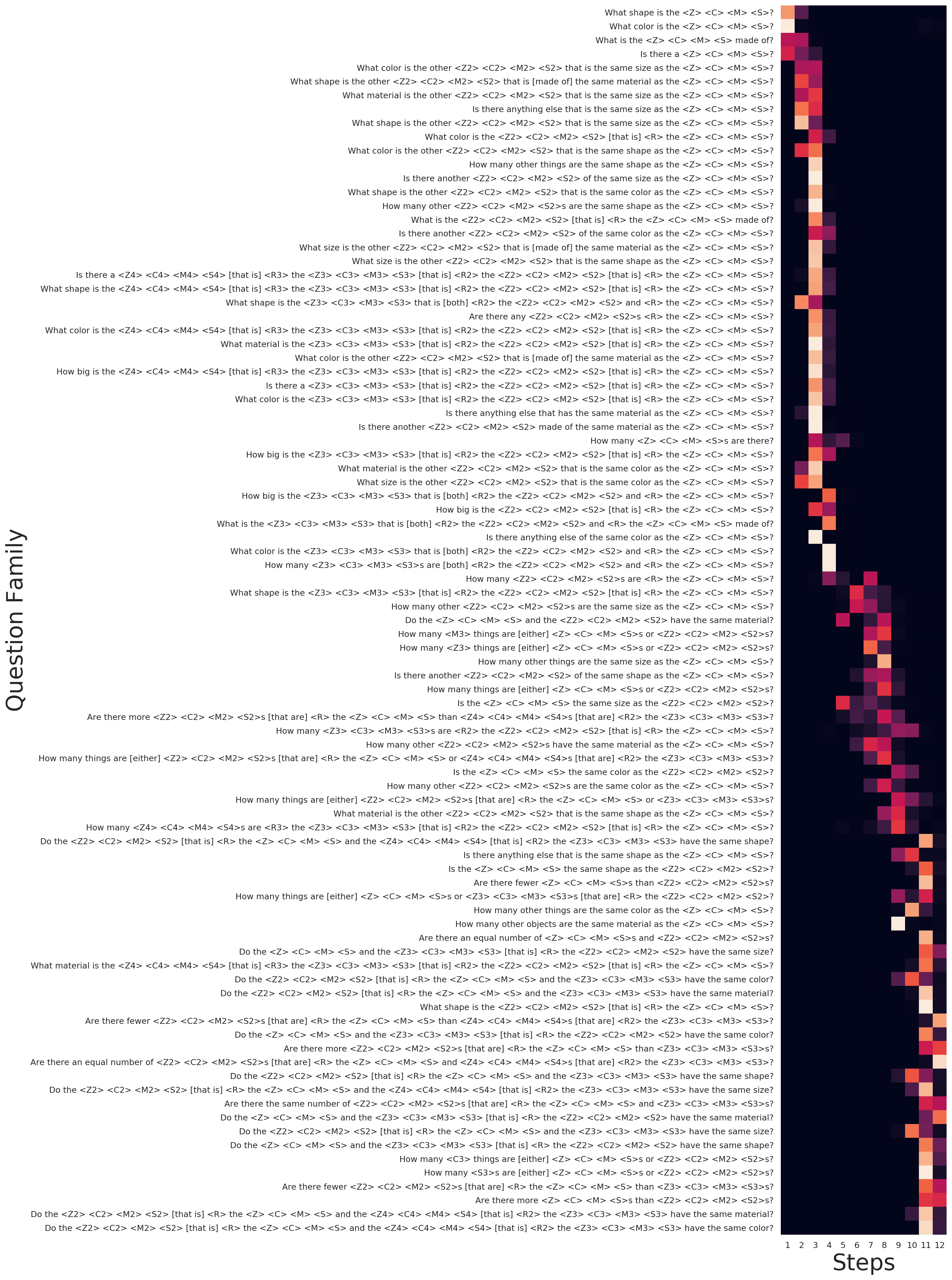}
  \end{center}
  \caption{
    The image above shows the average number of steps used by DACT-MAC ($\lambda = \num{5e-3}$) for each of the question families present in CLEVR, along with one template for each family to typify the whole group.
    Following \cite{clevr}, each question is generated by replacing \textit{$<$Z$>$, $<$C$>$, $<$M$>$, and $<$S$>$} with the size, color, material and/or shape of objects present in the image.
    Note that families with fewer supporting objects are more likely to be answered in less steps, and finding the number of objects that possess a pair of qualities (\textit{[both]}) is regarded as generally easier than finding those that possess \textit{[either]}.
    }
\label{fig:corr_full}
\end{figure*}

For any given synthetic image in the CLEVR dataset, a series of queries are generated by chaining a sequence of modular operations such as \textit{count}, \textit{filter}, \textit{compare}, \etc.
These functional programs can then be expressed in natural language in multiple ways, for instance translating $count(filter color(red, scene()))$ into \textit{“How many $<$C$>$ $<$M$>$ things are there?”}, a translation which is accomplished by instantiating the text templates specific to each program following \cite{clevr}.
As a result, questions with the same functional program can be clustered together into question families that share a similar complexity.
Figure \ref{fig:corr_full} includes a text template for each of the question families present in CLEVR, sorted by the average number of steps used for validation questions belonging to the specific family.

\subsection{Proofs}

In this section we prove that our method for building the final answer $Y$ can be interpreted as attending the \textit{intermediate outputs} $y_n$, with attention weights that follow a valid probability distribution.
We include two proofs by induction to show that, for any $n$, the accumulated answer $a_n$ can be expressed as a weighted sum of all \textit{intermediate outputs} up to the $n$th step, and that these weights always add up to one.

\noindent \textit{Proposition.} Every accumulated answer $a_n$ can be expressed as a weighted sum of all \textit{intermediate outputs} up to the $n$th step.
\begin{proof}
  Assume $\alpha_i$ exists for each $y_i$ such that every $a_{n-1}=y_{n-1} \alpha_{n-1} + \dots + y_0 \alpha_0$.
  This is trivial to prove for $n=1$ as $p_0 = 1$ makes $a_1 = y_1 p_0 + a_0 ( 1 - p_0 )$ become $a_1 = y_1$.
  \begin{align*}
    a_n &= y_n p_{n-1} + a_{n-1} ( 1 - p_{n-1} )\\
    &= y_n p_{n-1} + (\alpha_{n-1} y_{n-1} + \dots + \alpha_0 y_0) ( 1 - p_{n-1} ) \\ 
    &= y_n p_{n-1} + \sum_{i=0}^{n-1} y_{i} (\alpha_{i} ( 1 - p_{n-1} )) &&\qedhere
  \end{align*}
\end{proof}

\noindent \textit{Proposition.} Every accumulated answer $a_n$ can be expressed as a weighted sum of all \textit{intermediate outputs} up to the $n$th step, and the sum of the weights is equal to one.
\begin{proof}
  The base case is again trivial to prove since $p_0 = 1$ when $n=1$.
  Using the proof above we define $\beta_i$ to be the weights used to express $a_n$ as a weighed sum of $y_i$ $\forall i \in [1, n]$.
  \begin{align*}
  \beta_i &= \begin{cases}
    p_{n-1} \qquad \text{if} \quad i=n\\
    \alpha_i (1 - p_{n-1}) \quad \text{otherwise}
  \end{cases}\\
  \end{align*}
  Assume $\alpha_i$ exists for each $y_i$ such that every $a_{n-1}=\alpha_{n-1} y_{n-1} + \dots + \alpha_0 y_0$ and $\sum_{i=0}^{n-1} \alpha_i= 1$.
  \begin{align*}
    \sum_{i=0}^n \beta_i &= p_{n-1} + \sum_{i=0}^{n-1} \alpha_i (1 - p_{n-1}) \\
    &= p_{n-1} + \sum_{i=0}^{n-1} \alpha_i - p_{n-1} \sum_{i=0}^{n-1} \alpha_i \\
    &= p_{n-1} + 1 - p_{n-1} \\
    &= 1 &&\qedhere
  \end{align*}
\end{proof}

\newpage


{\small
\bibliographystyle{ieee_fullname}
\bibliography{PAPER.bbl}
}

\end{document}